\documentclass[conference]{IEEEtran}
\usepackage{cite}
\usepackage{amsmath,amssymb,amsfonts}
\usepackage{algorithmic}
\usepackage{graphicx}
\usepackage{textcomp}
\usepackage{xcolor}
\usepackage{url}
\def\BibTeX{{\rm B\kern-.05em{\sc i\kern-.025em b}\kern-.08em
    T\kern-.1667em\lower.7ex\hbox{E}\kern-.125emX}}

\usepackage{siunitx}
\usepackage{booktabs,colortbl, array}
\usepackage{pgfplotstable}
\pgfplotsset{compat=1.8}

\usepackage[caption=false]{subfig}

\newcommand\copyrighttext{%
\footnotesize \copyright 2019 IEEE. Personal use of this material is permitted.  Permission from IEEE must be obtained for all other uses, in any current or future media, including reprinting/republishing this material for advertising or promotional purposes, creating new collective works, for resale or redistribution to servers or lists, or reuse of any copyrighted component of this work in other works.\\
\url{https://doi.org/10.1109/FAS-W.2019.00057}
}
\newcommand\copyrightnotice{%
	\begin{tikzpicture}[remember picture,overlay]
		\node[anchor=south,yshift=0pt] at (current page.south) {\fbox{\parbox{\dimexpr\textwidth-\fboxsep-\fboxrule\relax}{\copyrighttext}}};
	\end{tikzpicture}%
}

\begin{document}

\title{Self-Organized Construction by Minimal Surprise}

\author{\IEEEauthorblockN{Tanja Katharina Kaiser}
\IEEEauthorblockA{
\textit{Institute of Computer Engineering}\\
\textit{University of L\"{u}beck}\\
L\"{u}beck, Germany \\
kaiser@iti.uni-luebeck.de}
\and
\IEEEauthorblockN{Heiko Hamann}
\IEEEauthorblockA{
\textit{Institute of Computer Engineering}\\
\textit{University of L\"{u}beck}\\
L\"{u}beck, Germany \\
hamann@iti.uni-luebeck.de}
}

\maketitle
\copyrightnotice

\begin{abstract}
  For the robots to achieve a desired behavior, we can program them directly, train them, or give them an innate driver that makes the robots themselves desire the targeted behavior. 
  With the minimal surprise approach, we implant in our robots the desire to make their world predictable. 
  Here, we apply minimal surprise to collective construction. Simulated robots push blocks in a 2D torus grid world.
  In two variants of our experiment we either allow for emergent behaviors or predefine the expected environment of the robots. 
  In either way, we evolve robot behaviors that move blocks to structure their environment and make it more predictable. 
  The resulting controllers can be applied in collective construction by robots.
\end{abstract}


\section{Introduction}
A simple approach to implement a robot swarm for self-organized construction or object aggregation is to define a probabilistic state machine: if there are few building blocks around, then pick up a block; if there are many blocks around, then drop yours~\cite{hamann2018}.
For more complex construction behaviors we would need to implement more complex state machines, test them, fix them, iterate, etc.
Here we follow a different approach. We deny the robots a defined task and instead implement an innate drive to prefer boring environments. 
The robots are free to develop behaviors that make sure to generate those boring environments.
We follow the minimal surprise approach which is roughly inspired by Friston~\cite{friston10}. 
In previous work, we have applied minimal surprise to generate collective behaviors, such as aggregation, flocking, and more recently self-assembly~\cite{kaiser18}.
We make the next step in complexity by providing the robots with building blocks in this scenario. 
These blocks can be pushed around by robots to form clusters of different sizes.
A~`boring environment' is then an environment with areas of few blocks and areas of many blocks.
Hence, the robots simplify to predict whether they may see a block. 

There are several approaches on collective construction with multiple cooperating robots that differ in their complexity. 
Some use sophisticated methods to calculate local rules for the robots offline~\cite{werfel14}. 
Others use simple reactive control but then also the construction behavior is limited to pushing and aggregating building materials~\cite{parker03,vardy18}.
Similarly for biological systems, there seem to be both variants, for example, sophisticated local rules in wasps~\cite{theraulazbonabeau95b} and blind bulldozing in ants~\cite{franks92b}. 
Here, we allow robots only to push blocks and the emerging behaviors aggregate blocks or form simple structures, such as lines.

Several approaches are similar to ours in terms of methodology, that is employing pairs of artificial neural networks (ANN). 
Ha and Schmidhuber~\cite{schmidhuber_2018} train world models and controllers (both are ANNs) separately for OpenAI Gym scenarios.
Generative Adversarial Nets (GANs) use an arms-race method to train artificial neural networks~\cite{goodfellow_generative_2014}. 
Turing learning is conceptually similar to GANs and was already applied to scenarios of swarm robotics~\cite{gross2017generalizing}.

The contribution of this paper is that we observe emergent robot controllers showing swarm construction behaviors. 
In our previous work~\cite{kaiser18}, we used a 2D~torus grid world with only robots living in it. 
Here, we increase the complexity of the environment by adding movable blocks. 
We need to change the sensor model, such that robots can detect other robots but also blocks and that they can distinguish them.
Before we had noticed that the emergence of interesting behaviors depends on the robot density (i.e., number of robots per area)~\cite{hamann14d,kaiser18}. 
Here we have to set the absolute robot and block densities but we also have to consider their ratio. 

\section{Methods}
\subsection{Experimental Setup}

In all of our experiments, we use a simulated homogeneous robot swarm of size~$N$ living on a 2D torus grid and we distribute $B$~blocks of building material in the environment. 
Each robot and each block occupy one grid cell. 
We use two different grid sizes~$L$ keeping the block density ($\frac{B}{L\times L}$) constant while varying the swarm density ($\frac{N}{L\times L}$).

\begin{figure}[t]
    \centering
     \subfloat[robot sensors]{\includegraphics[width=0.44\linewidth]{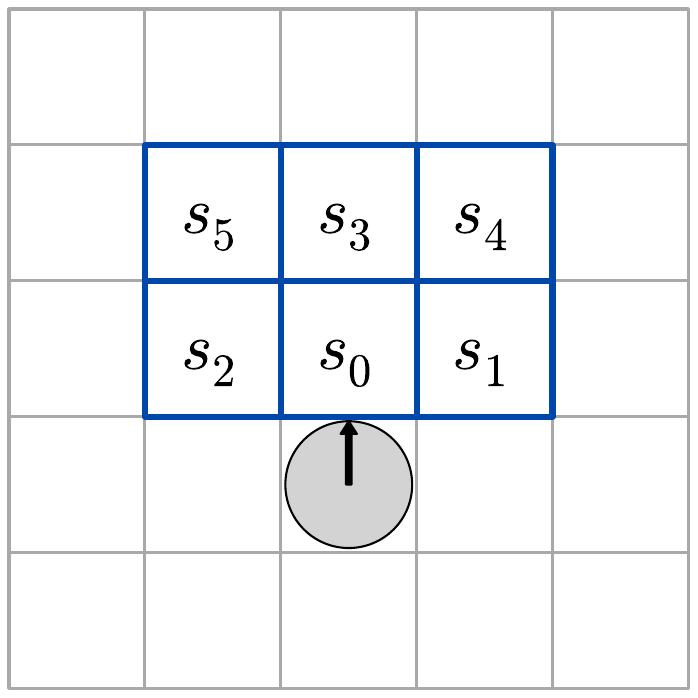}\label{fig:robotsensor}}
     \hspace{2mm}
     \subfloat[block sensors]{\includegraphics[width=0.44\linewidth]{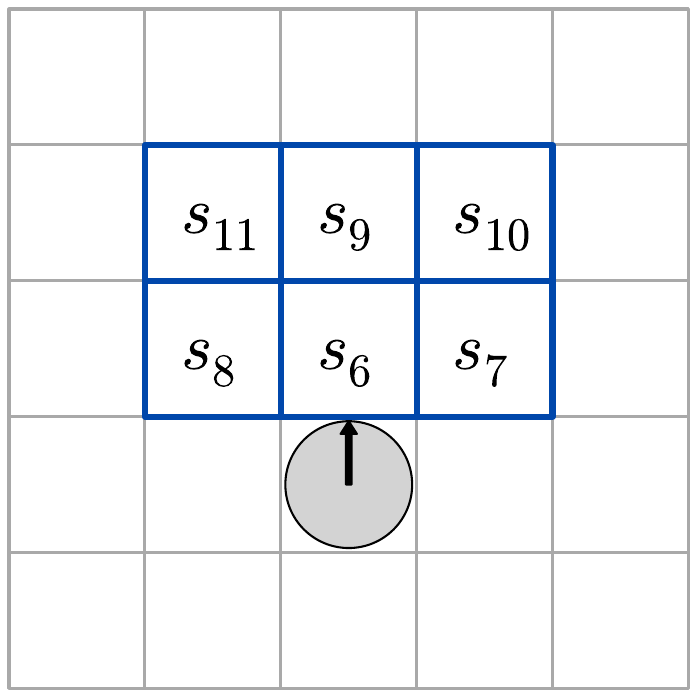}\label{fig:blocksensor}}
    \caption{Sensor model. The \textit{gray circle} represents the robot. The \textit{arrow} indicates its heading.}
\label{fig:sensormodel}
\end{figure}

The robots have discrete headings: North, South, East, and West. 
In each time step, robots can either move one grid cell forward or rotate on the spot. 
A~move forward is only possible if the grid cell in front is not occupied by another robot. 
A~robot pushes a single block one grid cell forward per time step if the block's target grid cell is empty. 
Each robot has two sets of binary sensors covering the six grid cells in front of it, see Fig.~\ref{fig:sensormodel}, that is, a total of 12~sensors. 
Sensors~$s_0, \dots, s_5$ (Fig.~\ref{fig:robotsensor}) sense robots while sensors $s_6, \dots, s_{11}$ (Fig.~\ref{fig:blocksensor}) observe blocks. 

\tikzset{%
  every neuron/.style={
    circle,
    draw,
    minimum size=0.7cm
  },
  neuron missing/.style={
    draw=none, 
    scale=2,
    text height=0.333cm,
    execute at begin node=\color{black}$\vdots$
  },
}

\begin{figure}[t]
    \centering
    \subfloat[action network]{
\resizebox {0.47\linewidth} {!} {
\begin{tikzpicture}[x=0.9cm, y=0.9cm, >=stealth,]
\foreach \m/\l [count=\y] in {1,missing,2,4}
  \node [every neuron/.try, neuron \m/.try] (input-\m) at (0,2 -\y*1.1) {};
\foreach \m [count=\y] in {1,missing,2}
  \node [every neuron/.try, neuron \m/.try ] (hidden-\m) at (1.75,2-\y*1.25) {};
\foreach \m [count=\y] in {1, 2}
  \node [every neuron/.try, neuron \m/.try ] (output-\m) at (3.2,1-\y) {};
\draw [<-] (input-1) -- ++(-1.6,0)
    node [above, midway] {$s_{0}(t)$};
\draw [<-] (input-2) -- ++(-1.6,0)
    node [above, midway] {$s_{11}(t)$};
\draw [<-] (input-4) -- ++(-1.6,0)
    node [above, midway] {$A(t-1)$};
\draw [->] (output-1) -- ++(1.2,0)
    node [above, midway] {$A(t)$};
\draw [->] (output-2) -- ++(1.2,0)
    node [above, midway] {$T(t)$};
\foreach \i in {1,2,4}
  \foreach \j in {1,...,2}
    \draw [->] (input-\i) -- (hidden-\j);
\foreach \i in {1,...,2}
  \foreach \j in {1, 2}
    \draw [->] (hidden-\i) -- (output-\j);
\end{tikzpicture}\label{fig:actionANN} }
    }
    \subfloat[prediction network]{
    \resizebox {0.47\linewidth} {!} {
  \begin{tikzpicture}[x=0.9cm, y=0.9cm, >=stealth,]
\foreach \m/\l [count=\y] in {1,missing,2,4}
  \node [every neuron/.try, neuron \m/.try] (input-\m) at (0,2 -\y*1.1) {};
\foreach \m [count=\y] in {1,missing,2}
  \node [every neuron/.try, neuron \m/.try ] (hidden-\m) at (1.75,1.4-\y) {};
\foreach \m [count=\y] in {1,missing,2}
  \node [every neuron/.try, neuron \m/.try ] (output-\m) at (3.5,1.4-\y) {};
\draw [<-] (input-1) -- ++(-1.6,0)
    node [above, midway] {$s_0(t)$};
\draw [<-] (input-2) -- ++(-1.6,0)
    node [above, midway] {$s_{11}(t)$};
\draw [<-] (input-4) -- ++(-1.6,0)
    node [above, midway] {$A(t)$};
\foreach \l [count=\i] in {0, 11}
  \draw [->] (output-\i) -- ++(2.1,0)
    node [above, midway] {$p_{\l}(t+1)$};
\foreach \i in {1,2,4}
  \foreach \j in {1,...,2}
    \draw [->] (input-\i) -- (hidden-\j);
\foreach \i in {1,...,2}
  \foreach \j in {1,...,2}
    \draw [->] (hidden-\i) -- (output-\j);
 \draw[->,shorten >=1pt] (hidden-1) to [out=90,in=140,loop,looseness=8.8] (hidden-1);
 \draw[->,shorten >=1pt] (hidden-2) to [out=270,in=220,loop,looseness=8.8] (hidden-2);
\end{tikzpicture}  }\label{fig:predictionANN}
    }
    \caption{Action network and prediction network. $A(t-1)$~is the robot's last action value and $A(t)$ is its next action. $T(t)$~is its turning direction. $s_0(t),\dots,s_{11}(t)$ are the robot's 12~sensor values at time step~t, ${p_0(t+1),\dots,p_{11}(t+1)}$ are its sensor predictions for time step $t+1$.}
    \label{fig:ANN}
\end{figure}
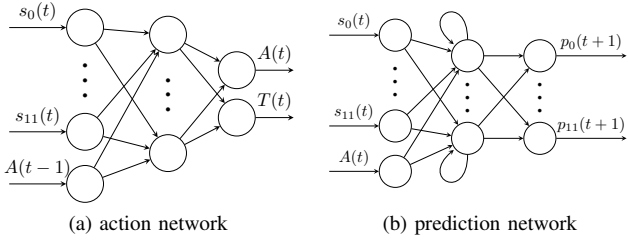

We equip each robot with a pair of ANN, see Fig.~\ref{fig:ANN}. 
The action network is implemented as a feedforward network and determines the robot's next action~$A(t)$.
It selects between straight motion and rotation and outputs a turning direction~$T(t)$ of $\pm 90^\circ$. 
The prediction network is a recurrent neural network enabling robots to predict their sensor values of the next time step. 
Both networks receive the robot's current sensor values as inputs. 
In addition, the action network receives the robot's last action~$A(t-1)$ and the prediction network its next action~$A(t)$. 

We evolve the ANNs in pairs using a simple genetic algorithm~\cite{holland75}. 
The genomes consist of two sets of weights that are randomly generated for the initial population: one for the action network and one for the prediction network. 
Each swarm member has an instance of the same genome in a given evaluation and thus, we use a homogeneous swarm. 
 
We reward correct sensor predictions. 
The prediction network receives direct selective pressure while the action network is solely subject to genetic drift. 
It receives pressure indirectly as it is paired with a pressured prediction network.
The fitness function is defined as 

\begin{equation}
F = \frac{1}{NTR} \sum_{t=0}^{T-1} \sum_{n=0}^{N-1} \sum_{r=0}^{R-1} 1 -  | p_{r}^{n}(t) - s_{r}^{n}(t) | \, ,
\label{equ:fitness}
\end{equation}

where $N$ is swarm size, $R$~is the number of sensors per robot, $p_{r}^{n}(t)$ is the prediction for sensor~$r$ of robot~$n$ at time step~$t$, and~$s_{r}^{n}(t)$~is the value of sensor~$r$. 

We run the evolutionary algorithm for $100$~generations and evaluate each genome in ten independent simulation runs (swarm size between 10 and~50) for $1000$~time steps each using random initial robot and block positions. 
The fitness of a genome is the minimum fitness~(Eq.~\ref{equ:fitness}) observed in those ten evaluations. 
For the evolutionary algorithm, we use a population size of~$50$, proportionate selection, elitism of one, and a mutation rate of~$0.1$ for both networks. 
We evaluate all scenarios in 20~independent evolutionary runs. 
Table~\ref{tab:TAB3} summarizes the used parameters. 

\begin{table}[t]
\caption{Parameter settings.\label{tab:TAB3}}
\begin{center}
\begin{tabular}{|l|l|}
\hline
 \textbf{parameter} & \textbf{value} \\ \hline 
 grid side length $L$ & \{16, 20\} \\ 
 \# of sensors $R$& 12 \\ 
 swarm size $N$ &  10 to 50 \\ 
 \# of blocks $B$ & \{32, 50, 75\}\\ \hline 
 population size & 50 \\ 
 number of generations & 100 \\ 
 evaluation length in time steps $T$ & 1000 \\ 
 \# of sim. runs per fitness evaluation & 10 \\  
 elitism & 1 \\ 
 mutation rate & 0.1 \\   \hline
\end{tabular}
\end{center}
\end{table}

\subsection{Metrics} \label{sec:metrics}
To validate our approach, we do several post-evaluations.
For the best evolved individuals, we measure the fitness (Eq.~\ref{equ:fitness}) and classify formed block structures based on predefined metrics. 
We determine the runs with altered block positions and assess the similarity of their block positions at the start and the end of the run as well as the movement of blocks and robots.   

The similarity of the block positions is the quantity of grid cells that were occupied by blocks at the start and that are still (or again) occupied by blocks at the end of the evaluation normalized by the total number of blocks (post-evaluation of the best evolved individual). 
It serves as an indicator to assess roughly how many blocks were moved by robots.

In addition, we measure the movement~$M$ of robots and blocks, respectively, over a time period of $\tau = \frac{L \times L}{2}$ time steps as by Hamann~\cite{hamann14d}. 
It represents the mean covered distance. 
We define the movement~$M$ as
\begin{equation}
	M = M_{x} + M_{y}\, ,
	\label{equ:movement}
\end{equation} 
where $M_{x}$ and $M_{y}$ are the movement in x- and y-direction on the grid. 
We define $M_{x}$ as  
\begin{equation}
M_{x} = \frac{1}{P\tau} \sum_{p=0}^{P-1} \sum_{t = T - \tau}^{T-1}| x_{p}(t) - x_{p}(t+1) |\, ,
\label{equ:dx}
\end{equation} 
and $M_{y}$ accordingly. 
In the case of measuring robot movement, $P$~is the swarm size~$P=N$ and in the case of measuring block movement it is the number of blocks~$P=B$. 

\begin{table*}[t]
\caption{Metrics (cf. Sec.~\ref{sec:metrics}) of 20~independent evolutionary runs. Median values in brackets. \label{tab:measurements}}
\begin{center}
\begin{tabular}{|c|c|c|c|c||c|c|c|c||c|c|c|c|c|}
\hline
\textbf{}& & & \textbf{grid} &\textbf{median} & \multicolumn{4}{c||}{\textbf{similarity $< 1.0$}} & & \multicolumn{4}{c|}{\textbf{structures}} \\
\cline{6-9}
\cline{11-14}
\textbf{robots} & \textbf{blocks} & \textbf{ratio} & \textbf{size} & \textbf{fitness} & \textbf{qty.}& \textbf{similarity} & \textbf{block mov.} & \textbf{robot mov.} & & \textbf{lines} & \textbf{pairs} & \textbf{clusters} & \textbf{disp.}\\
\hline
10 & 32 & $5:16$ & $16\times 16$ & 0.91 & 11 & \begin{tabular}{@{}c@{}}0.778 \\ (0.875) \end{tabular} & \begin{tabular}{@{}c@{}}0.0 \\ (0.0) \end{tabular} &  \begin{tabular}{@{}c@{}}0.43 \\ (0.47) \end{tabular} & \begin{tabular}{@{}c@{}}start \\ end\end{tabular} & \begin{tabular}{@{}c@{}} 0.0 \\ 0.0 \end{tabular} & \begin{tabular}{@{}c@{}} 20.0 \\ 27.5 \end{tabular} & \begin{tabular}{@{}c@{}} 0.0\\ 5.0 \end{tabular} & \begin{tabular}{@{}c@{}} 80.0 \\ 67.5 \end{tabular}  \\ \hline
16 & 32 & $1:2$ & $16\times 16$ & 0.898 & 11 & \begin{tabular}{@{}c@{}}0.648 \\ (0.781)\end{tabular} & \begin{tabular}{@{}c@{}}0.0 \\ (0.0) \end{tabular} & \begin{tabular}{@{}c@{}}0.49 \\ (0.48) \end{tabular} & \begin{tabular}{@{}c@{}}start \\ end\end{tabular} & \begin{tabular}{@{}c@{}} 0.0 \\ 2.5  \end{tabular} & \begin{tabular}{@{}c@{}} 15.0 \\ 15.0 \end{tabular} & \begin{tabular}{@{}c@{}} 0.0 \\ 12.5 \end{tabular} & \begin{tabular}{@{}c@{}} 85.0 \\ 70.0 \end{tabular} \\ \hline 
32 & 32 & $1:1$ & $16\times 16$ & 0.892 & 14 & \begin{tabular}{@{}c@{}}0.418 \\ (0.344) \end{tabular} & \begin{tabular}{@{}c@{}}0.0 \\ (0.0) \end{tabular} & \begin{tabular}{@{}c@{}}0.38 \\ (0.42) \end{tabular} & \begin{tabular}{@{}c@{}}start \\ end\end{tabular} & \begin{tabular}{@{}c@{}} 0.0 \\ 5.0 \end{tabular} & \begin{tabular}{@{}c@{}} 7.5 \\ 47.5  \end{tabular} & \begin{tabular}{@{}c@{}} 0.0 \\ 10.0 \end{tabular} & \begin{tabular}{@{}c@{}} 92.5 \\ 37.5 \end{tabular}\\ \hline 
20 & 50 & $2:5$ & $20\times 20$ & 0.902 & 10 & \begin{tabular}{@{}c@{}}0.848 \\ (0.870) \end{tabular} & \begin{tabular}{@{}c@{}}0.0 \\ (0.0) \end{tabular} & \begin{tabular}{@{}c@{}}0.33 \\ (0.42) \end{tabular} & \begin{tabular}{@{}c@{}}start \\ end\end{tabular} & \begin{tabular}{@{}c@{}} 0.0 \\ 0.0 \end{tabular} & \begin{tabular}{@{}c@{}} 2.5 \\ 7.5 \end{tabular} & \begin{tabular}{@{}c@{}} 0.0 \\ 0.0 \end{tabular} & \begin{tabular}{@{}c@{}} 97.5 \\ 92.5 \end{tabular} \\ \hline 
25 & 50 & $1:2$ & $20\times 20$ & 0.903 & 10 & \begin{tabular}{@{}c@{}}0.816 \\ (0.830) \end{tabular} & \begin{tabular}{@{}c@{}}0.0 \\ (0.0) \end{tabular} & \begin{tabular}{@{}c@{}}0.43 \\ (0.45) \end{tabular} & \begin{tabular}{@{}c@{}}start \\ end\end{tabular} & \begin{tabular}{@{}c@{}} 0.0 \\ 0.0 \end{tabular} & \begin{tabular}{@{}c@{}} 7.5\\ 12.5 \end{tabular} & \begin{tabular}{@{}c@{}}0.0 \\ 0.0 \end{tabular} & \begin{tabular}{@{}c@{}} 92.5 \\ 87.5  \end{tabular}\\ \hline 
50  & 50 & $1:1$ & $20\times 20$ & 0.866 & 7 & \begin{tabular}{@{}c@{}}0.494 \\ (0.420) \end{tabular} & \begin{tabular}{@{}c@{}}0.0 \\ (0.0) \end{tabular} & \begin{tabular}{@{}c@{}}0.24 \\ (0.29) \end{tabular} &\begin{tabular}{@{}c@{}}start \\ end\end{tabular}& \begin{tabular}{@{}c@{}} 0.0 \\ 5.0 \end{tabular} & \begin{tabular}{@{}c@{}} 12.5 \\ 27.5 \end{tabular} & \begin{tabular}{@{}c@{}}0.0 \\  0.0\end{tabular} & \begin{tabular}{@{}c@{}} 87.5\\ 67.5 \end{tabular} \\ \hline 
25 & 75 & $1:3$ & $20\times 20$ & 0.862 & 6 & \begin{tabular}{@{}c@{}} 0.780 \\ (0.820) \end{tabular} & \begin{tabular}{@{}c@{}}0.0 \\ (0.0) \end{tabular} & \begin{tabular}{@{}c@{}}0.20 \\ (0.20) \end{tabular} & \begin{tabular}{@{}c@{}}start \\ end\end{tabular} & \begin{tabular}{@{}c@{}} 5.0 \\ 15.0 \end{tabular} & \begin{tabular}{@{}c@{}} 50.0 \\ 47.5  \end{tabular} & \begin{tabular}{@{}c@{}} 0.0 \\ 0.0  \end{tabular} & \begin{tabular}{@{}c@{}} 45.0 \\ 37.5  \end{tabular} \\ 
\hline
\end{tabular}
\end{center}
\end{table*}

We classify formed block structures as pairs, lines, clusters, or dispersion based on their highest resemblance (automated using Python scripts) at the start and the end of the evaluation of the best evolved individual.
\textit{Pairs} and \textit{lines} are formed by blocks placed next to each other horizontally or vertically and differ only in length.
Pairs consist of two blocks while lines are at least three blocks long. 
Both structures can have up to half of their length of neighbors on each side next to them, whereby no two adjacent grid cells parallel to the structure are allowed to be occupied by blocks. 
\textit{Clusters} are blocks that have at least four blocks in their Moore neighborhoods and at least three in their von Neumann neighborhood.
Blocks classified as lines cannot be part of a cluster. 
\textit{Dispersed} blocks have maximally one direct diagonal neighbor. 

\section{Results}

\subsection{Impact of the Robot-to-Block Ratio}

In the following experiments, we use a constant block density of $12.5\%$ by setting either 32~blocks on a $16\times 16$ grid or 50~blocks on a $20\times 20$ grid. 
On both grids, we do experiments with robot-to-block ratios of~$1:1$ (high swarm density) and~$1:2$ (intermediate swarm density). 
In addition, we use a ratio of~$5:16$ on the smaller grid and of~$2:5$ on the larger grid (low swarm densities), cf. Table~\ref{tab:measurements}. 
We increase the block density to $18.75\%$ in one experiment to show the effects on the resulting structures. 

\begin{figure}[t]
    \centering
    \includegraphics[width=0.9\linewidth]{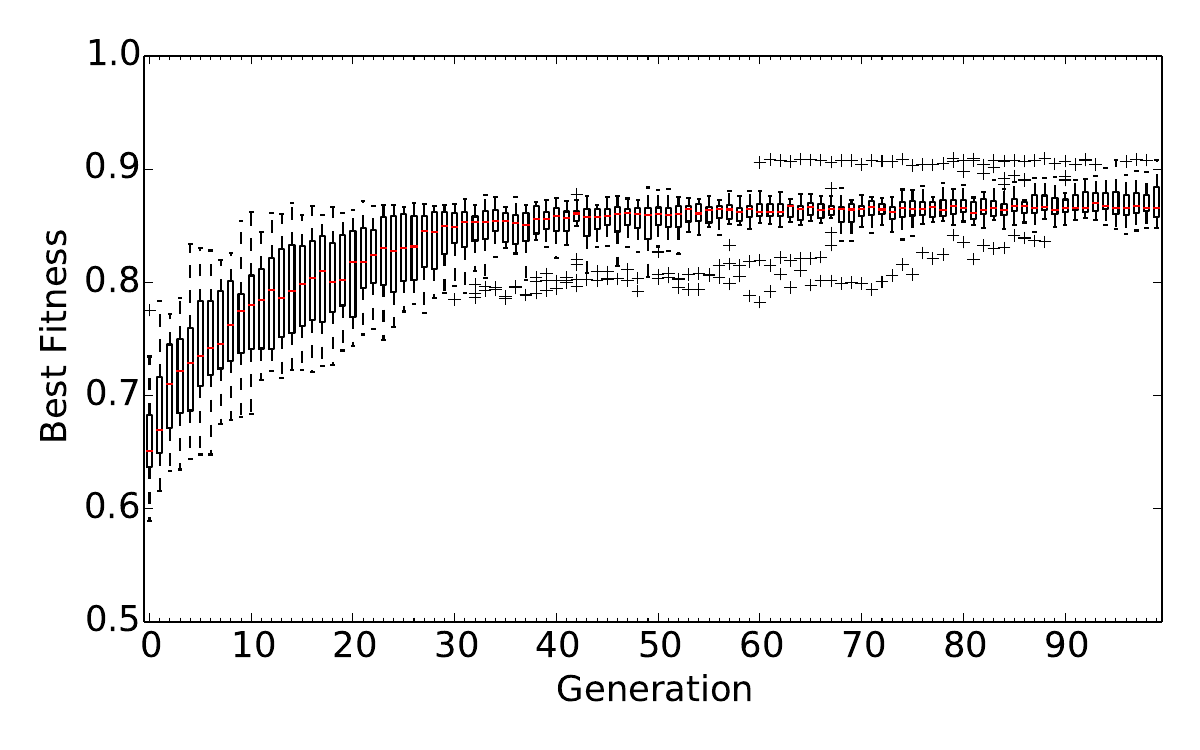}
    \caption{Best fitness of 20~independent evolutionary runs on the $20\times 20$ grid with 50~robots and 50~blocks. Medians are indicated by the red bars.\label{fig:fitness}}
\end{figure}

\begin{figure*}[t]
    \centering
    \subfloat[Clusters (10 robots, 32 blocks)]{\includegraphics[width=0.22\linewidth]{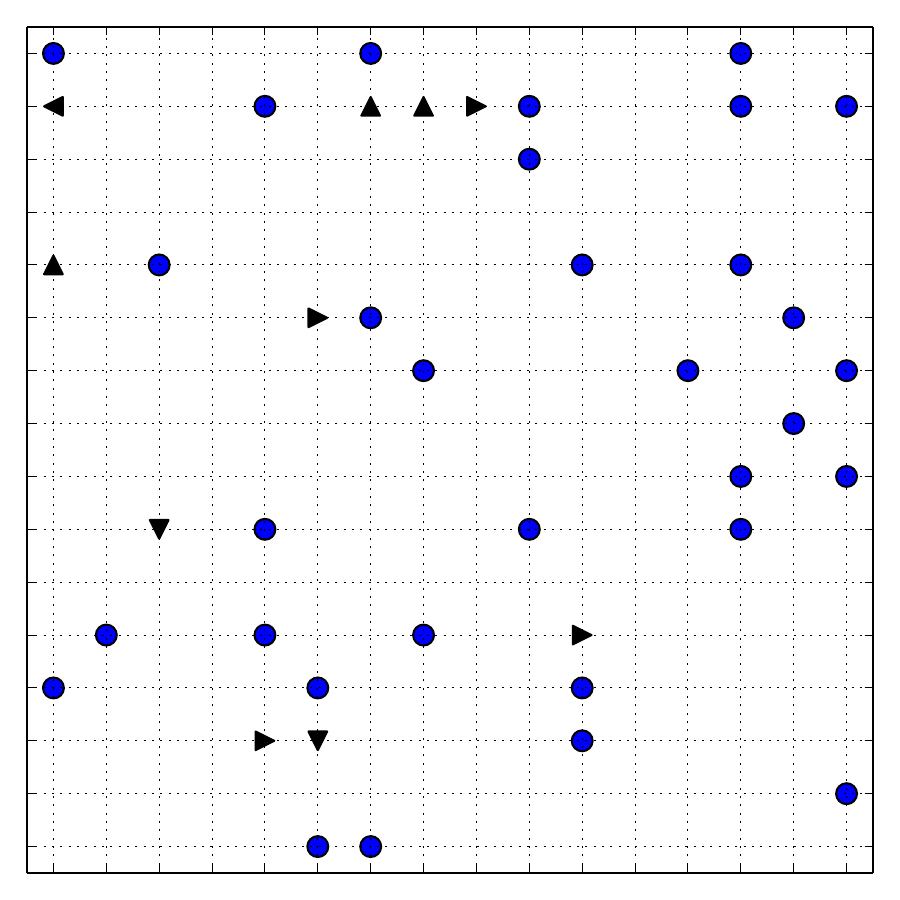}\includegraphics[width=0.22\linewidth]{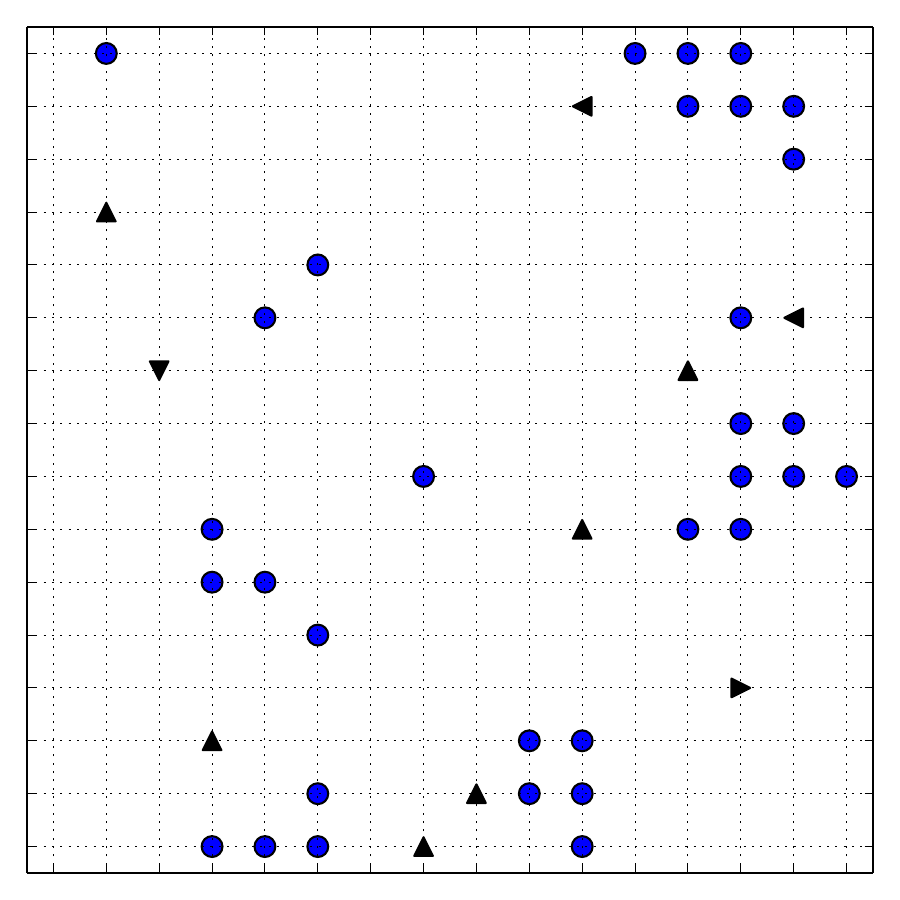}\label{fig:cluster}} \hspace{2mm}
     \subfloat[Pairs (20 robots, 50 blocks)]{\includegraphics[width=0.22\linewidth]{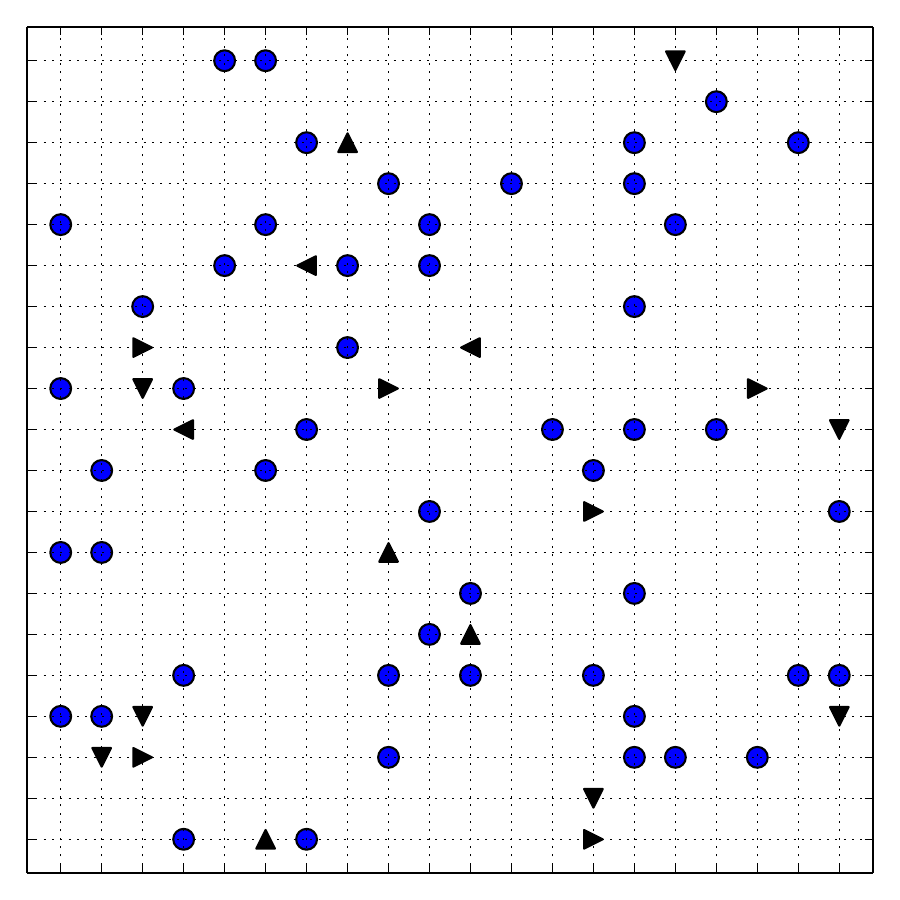}\includegraphics[width=0.22\linewidth]{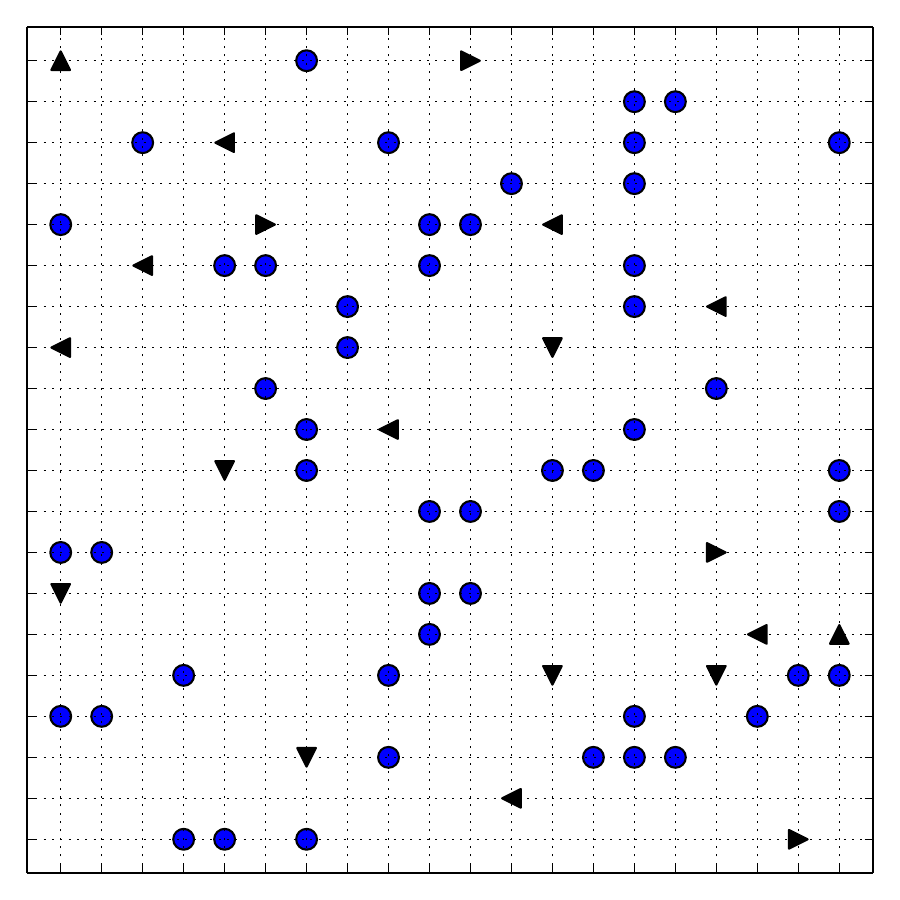}\label{fig:pairs}} \\
     \subfloat[Dispersion (16 robots, 32 blocks)]{\includegraphics[width=0.22\linewidth]{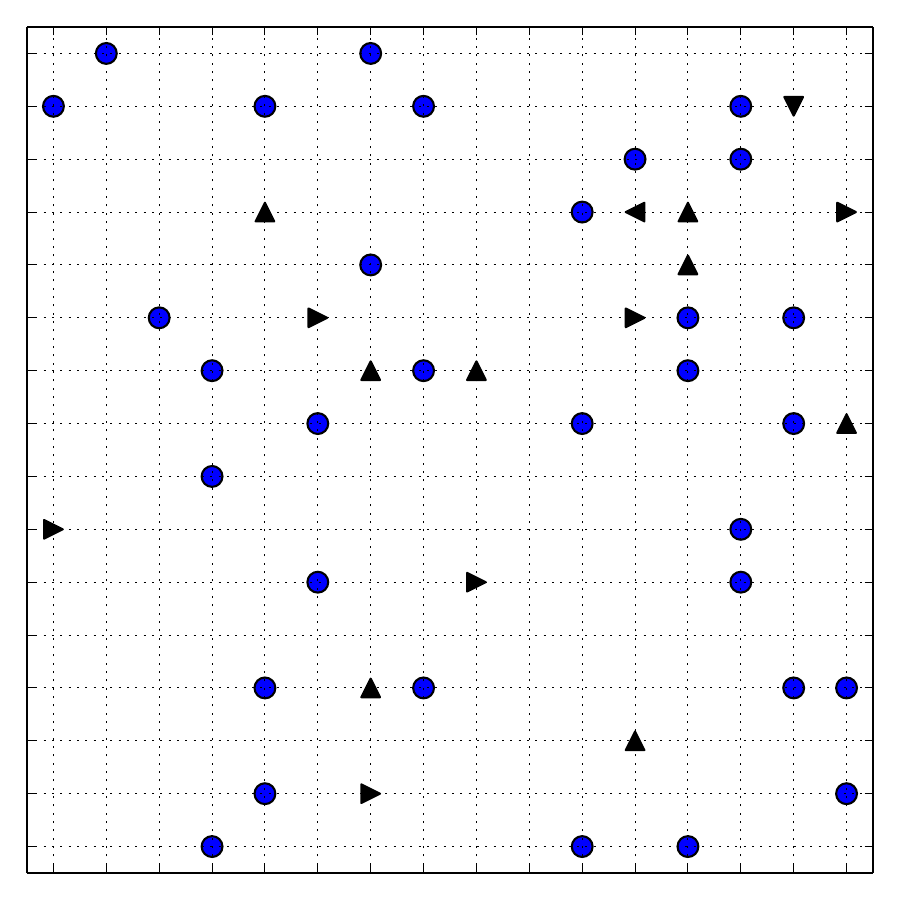}\includegraphics[width=0.22\linewidth]{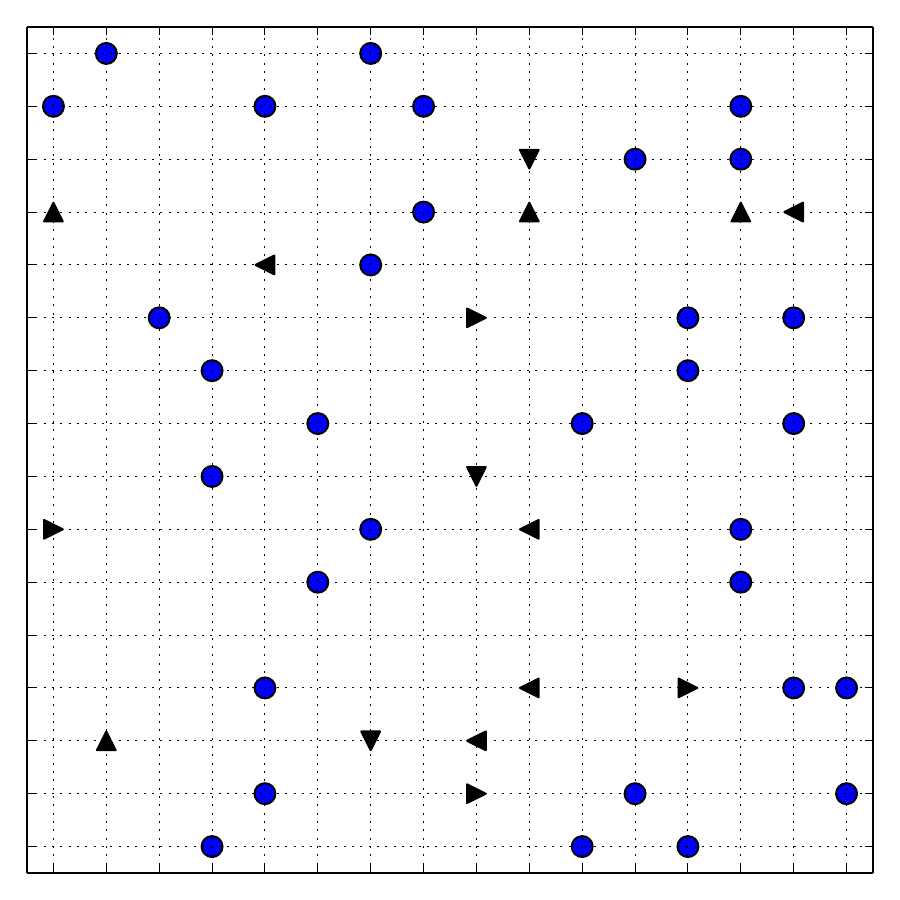}\label{fig:disp}} \hspace{2mm}
     \subfloat[Lines (25 robots, 75 blocks)]{\includegraphics[width=0.22\linewidth]{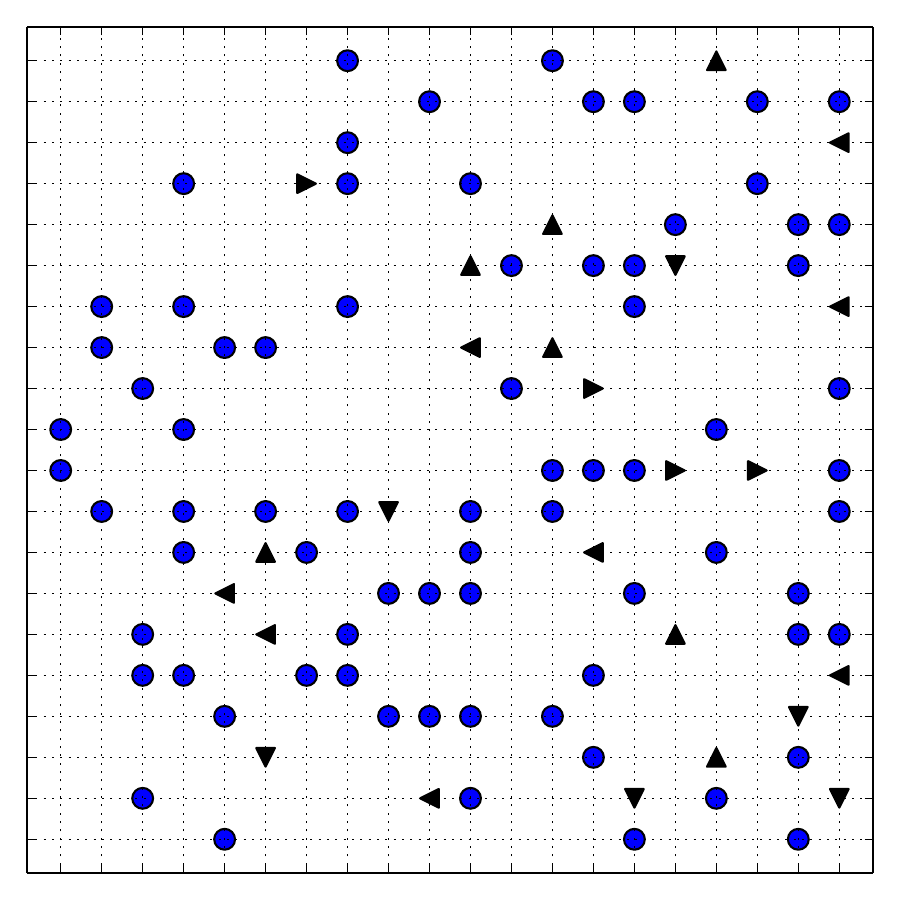}\includegraphics[width=0.22\linewidth]{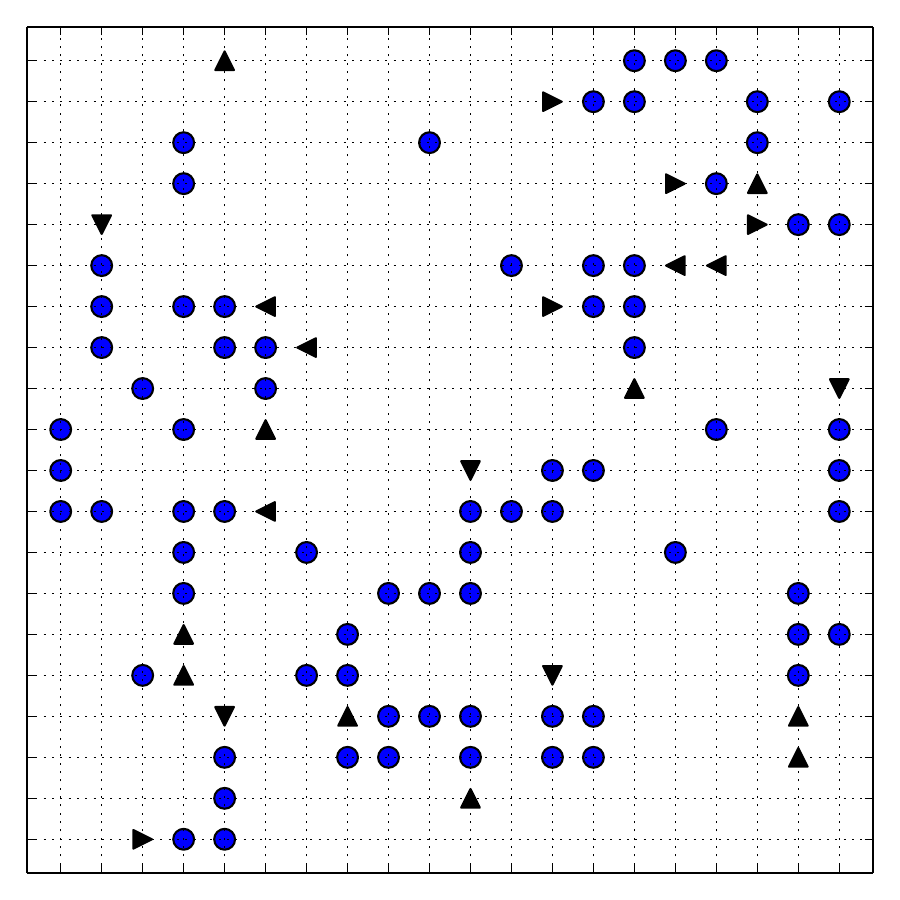}\label{fig:lines75}}
    \caption{Structures at the start (left) and end (rights) of a run. Robots are represented by triangles, blocks by circles. Triangles give the robots' headings.}
\label{fig:structures}
\end{figure*}

A median best fitness of at least $0.86$ in the last generation is reached in all experiments, see Table~\ref{tab:measurements}, meaning that $86\%$ and more of the sensor values are assessed correctly by the prediction networks. 
Figure~\ref{fig:fitness} shows the increase of the best fitness over generations of 20~independent evolutionary runs using 50~robots and 50~blocks on a $20\times 20$ grid. 
It is representative for the fitness curves observed in all experiments. 

We compare the quantity of runs with altered block positions, that is, with a similarity lower than~$1.0$. 
For the smaller robot-to-block ratios, half of the runs on the larger grid and $11$ runs on the smaller grid lead to the alteration of block positions. 
The number rises to $14$ on the smaller grid and decreases to seven on the larger grid for a $1:1$ ratio. 

As expected, we find that increasing robot-to-block ratios lead to decreasing similarities (i.e., more moved blocks). 
Precisely, runs with a $1:1$ ratio have about $40$~percentage points~(pp) lower similarities than smaller ratios and thus, block positions are altered most.

For all runs, we measured no block movement during the last~$\tau$~time steps (i.e., $\tau=128$ for the $16\times 16$~grid, $\tau = 200$ for the $20\times 20$~grid). 
Thus, the system converged as blocks have fixed positions and form a stable structure. 
In all runs with a similarity below $1.0$, we find robot movement. 
In contrast, the robot movement is mostly zero in runs without altered block positions indicating that robots mostly turn. 
We infer that the prediction task is easy as high fitness values are reached in all runs. 
Nevertheless, in a few runs robots move constantly without pushing any blocks or self-assemble into structures. 
Thus, we find a variety of swarm behaviors emerging due to our task-independent reward. 

We compare the formed block structures at the start and the end of the runs to estimate how much was changed by the robots.
A video of resulting behaviors can be found online.\footnote{https://youtu.be/T9s5669ypXM} 
The random initialization of block positions is mostly classified as dispersion and in less than $20\%$ of the runs as pairs except for the scenario with a $1:3$~robot-to-block ratio, cf. Tab.~\ref{tab:measurements}. 
Denser block distributions reduce the probability of dispersion and increase the probability of pairs. 
For the lower robot-to-block ratios, the best evolved behaviors decrease dispersion by $5$ to $15$~pp while they decrease dispersion by $55$~pp on the smaller grid and by $20$~pp on the larger grid for the $1:1$ ratio. 
Thus, we find that robots do not push blocks into a new structure in all runs with altered block positions. 
Scenarios with lower robot-to-block ratios lead to only one to three runs with modified structures out of ten to $11$ runs with altered block positions while scenarios with a $1:1$ ratio lead to four out of seven runs on the larger grid and $11$ out of $14$ on the smaller grid. 
Thus, scenarios with a $1:1$ ratio lead to the greatest alteration of block structures. 

We find that dispersion (Fig.~\ref{fig:disp}) is the most frequent structure in all experiments except for the $1:3$ robot-to-block ratio case. 
Pairs (Fig.~\ref{fig:pairs}) form frequently in all experiments while lines emerge rarely.
Only in the $1:3$ ratio case, the amount of pairs decreases during the run.
Clusters (Fig.~\ref{fig:cluster}) form only on the smaller grid, maybe because robots need to push blocks more grid cells forward to group blocks on the larger grid. 

In summary, we find that a $1:1$ robot-to-block ratio results in the most active swarm construction behaviors that change the initial block distribution a lot by forming structures. 
Different structures form on the two grid sizes. 
Using a higher block density with a $1:3$ robot-to-block ratio did not improve our results and thus, we focus on the first six experiments in the following.  

\subsection{Engineered Self-Organized Construction}

\begin{table*}[t]
\caption{Metrics (cf. Sec.~\ref{sec:metrics}) of 20~independent evolutionary runs with predefined predicted pairs. Median values in brackets. \label{tab:measurementsPREL}}
\begin{center}
\begin{tabular}{|c|c|c|c|c||c|c|c||c|c|c|c|c|}
\hline
& & & \textbf{grid} & \textbf{median} & & \textbf{block} & \textbf{robot}& \multicolumn{1}{c|}{} &\multicolumn{4}{c|}{\textbf{structures}} \\
\cline{10-13} 
\textbf{robots} & \textbf{blocks} & \textbf{ratio} & \textbf{size} & \textbf{fitness} & \textbf{similarity (\%)}  & \textbf{movement} & \textbf{movement} & \multicolumn{1}{c|}{} & \textbf{lines} & \textbf{pairs} & \textbf{clusters} & \textbf{dispersion} \\
\hline
10 & 32 & $5:16$ & $16\times 16$ & 0.873 & \begin{tabular}{@{}c@{}}0.661 \\ (0.688)\end{tabular}  & \begin{tabular}{@{}c@{}} 0.03 \\ (0.03) \end{tabular} & \begin{tabular}{@{}c@{}} 0.25 \\ (0.24) \end{tabular} &  \begin{tabular}{@{}c@{}}start \\ end\end{tabular} & \begin{tabular}{@{}c@{}}0.0 \\ 0.0 \end{tabular} & \begin{tabular}{@{}c@{}}20.0 \\ 85.0\end{tabular} & \begin{tabular}{@{}c@{}}0.0 \\ 2.5\end{tabular} & \begin{tabular}{@{}c@{}}80.0 \\ 12.5\end{tabular} \\ \hline
16 & 32 & $1:2$ & $16\times 16$ & 0.854 & \begin{tabular}{@{}c@{}}0.559 \\ (0.562)\end{tabular} & \begin{tabular}{@{}c@{}} 0.02 \\ (0.0) \end{tabular} & \begin{tabular}{@{}c@{}} 0.21 \\ (0.22) \end{tabular} &  \begin{tabular}{@{}c@{}}start \\ end\end{tabular} & \begin{tabular}{@{}c@{}}0.0 \\ 10.0 \end{tabular}  & \begin{tabular}{@{}c@{}}12.5 \\ 72.5 \end{tabular}  & \begin{tabular}{@{}c@{}}0.0 \\ 5.0 \end{tabular}  & \begin{tabular}{@{}c@{}}87.5 \\ 12.5 \end{tabular}  \\ \hline 
32 & 32 & $1:1$ & $16\times 16$ & 0.808 & \begin{tabular}{@{}c@{}}0.439 \\ (0.422)\end{tabular} & \begin{tabular}{@{}c@{}} 0.0 \\ (0.0) \end{tabular} & \begin{tabular}{@{}c@{}} 0.26 \\ (0.26) \end{tabular} &\begin{tabular}{@{}c@{}}start \\ end\end{tabular} & \begin{tabular}{@{}c@{}}0.0 \\ 15.0 \end{tabular} & \begin{tabular}{@{}c@{}}12.5 \\ 55.0 \end{tabular}  & \begin{tabular}{@{}c@{}}0.0 \\ 15.0 \end{tabular}  & \begin{tabular}{@{}c@{}}87.5 \\ 15.0 \end{tabular}     \\ \hline 
20 & 50 & $2:5$ & $20\times 20$ & 0.868 & \begin{tabular}{@{}c@{}}0.552 \\ (0.560)\end{tabular} & \begin{tabular}{@{}c@{}} 0.01 \\ (0.0) \end{tabular} & \begin{tabular}{@{}c@{}} 0.18 \\ (0.20) \end{tabular} &\begin{tabular}{@{}c@{}}start \\ end\end{tabular} & \begin{tabular}{@{}c@{}}0.0 \\ 15.0 \end{tabular}  &  \begin{tabular}{@{}c@{}}0.0 \\ 75.0 \end{tabular} & \begin{tabular}{@{}c@{}}0.0 \\ 5.0 \end{tabular} &  \begin{tabular}{@{}c@{}}100.0 \\ 5.0 \end{tabular}  \\ \hline 
25 & 50 & $1:2$ & $20\times 20$ & 0.861 & \begin{tabular}{@{}c@{}}0.519 \\ (0.530)\end{tabular} & \begin{tabular}{@{}c@{}} 0.01 \\ (0.0) \end{tabular} & \begin{tabular}{@{}c@{}} 0.18 \\ (0.18) \end{tabular} &\begin{tabular}{@{}c@{}}start \\ end\end{tabular} & \begin{tabular}{@{}c@{}}0.0 \\ 5.0 \end{tabular}  & \begin{tabular}{@{}c@{}}7.5 \\ 85.0 \end{tabular}  & \begin{tabular}{@{}c@{}}0.0 \\ 5.0 \end{tabular}  & \begin{tabular}{@{}c@{}}92.5 \\ 5.0 \end{tabular}      \\ \hline 
50  & 50 & $1:1$ & $20\times 20$ & 0.817 & \begin{tabular}{@{}c@{}}0.421 \\ (0.430)\end{tabular} & \begin{tabular}{@{}c@{}} 0.0 \\ (0.0) \end{tabular} & \begin{tabular}{@{}c@{}} 0.29 \\ (0.30) \end{tabular} &\begin{tabular}{@{}c@{}}start \\ end\end{tabular} & \begin{tabular}{@{}c@{}}0.0 \\ 5.0 \end{tabular}  &  \begin{tabular}{@{}c@{}}7.5 \\ 55.0 \end{tabular} & \begin{tabular}{@{}c@{}}0.0 \\ 15.0 \end{tabular} &  \begin{tabular}{@{}c@{}}92.5 \\ 25.0 \end{tabular}   \\ \hline 
\end{tabular}
\end{center}
\end{table*}

\begin{figure}[t]
    \centering
    \includegraphics[width=0.9\linewidth]{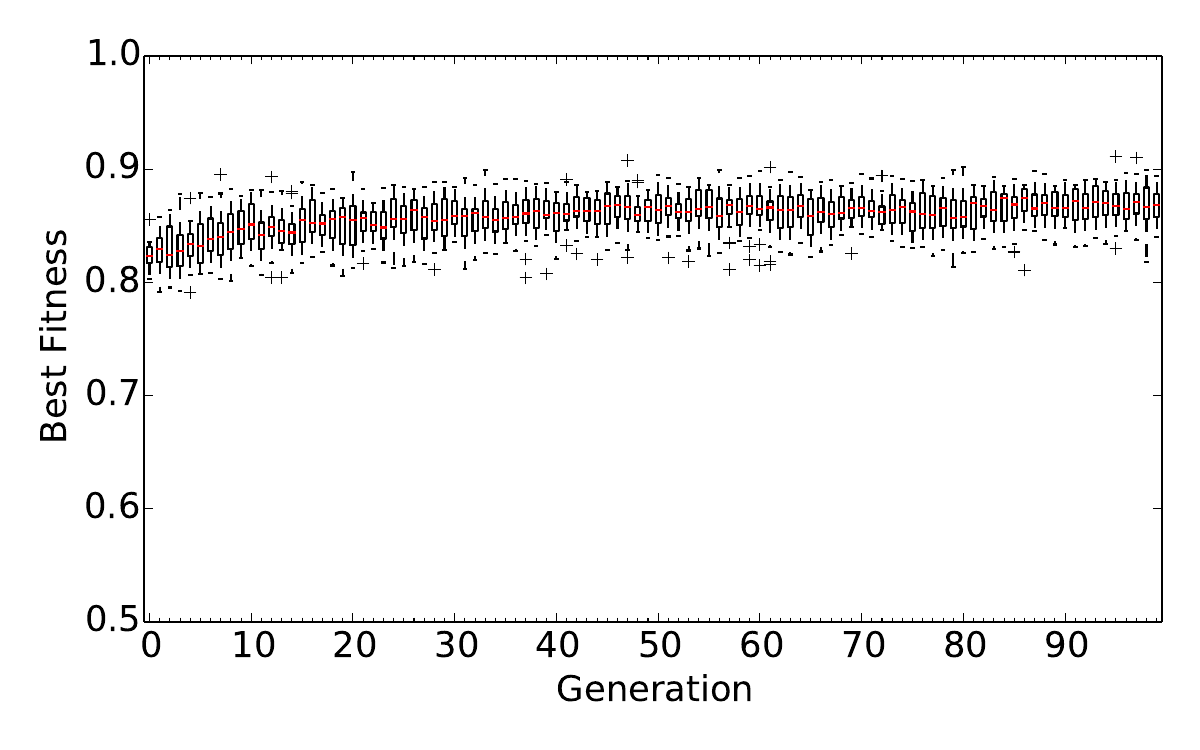}
    \caption{Best fitness of 20~independent runs on the $20\times 20$ grid with 20~robots and 50~blocks and predefined pairs. Medians are indicated by the red bars.}
\label{fig:fitnessPRE}
\end{figure}

\begin{table*}[t]
\caption{Metrics (cf. Sec.~\ref{sec:metrics}) of 20~independent evolutionary runs with predefined predicted clusters. Median values in brackets. \label{tab:measurementsPREC}}
\begin{center}
\begin{tabular}{|c|c|c|c|c||c|c|c||c|c|c|c|c|}
\hline
& & & \textbf{grid} & \textbf{median} & & \textbf{block} & \textbf{robot} &  \multicolumn{1}{c|}{} &\multicolumn{4}{c|}{\textbf{structures}} \\
\cline {10-13}
\textbf{robots} & \textbf{blocks} & \textbf{ratio} & \textbf{size} & \textbf{fitness} & \textbf{similarity (\%)}  & \textbf{movement} & \textbf{movement} & \multicolumn{1}{c|}{} & \textbf{lines} & \textbf{pairs} & \textbf{clusters} & \textbf{dispersion} \\
\hline
10 & 32 & $5:16$ & $16\times 16$ & 0.658 & \begin{tabular}{@{}c@{}}0.567 \\ (0.546)\end{tabular} & \begin{tabular}{@{}c@{}} 0.05 \\ (0.06) \end{tabular} & \begin{tabular}{@{}c@{}} 0.24 \\ (0.29) \end{tabular} &  \begin{tabular}{@{}c@{}}start \\ end\end{tabular} & \begin{tabular}{@{}c@{}} 0.0 \\ 2.5 \end{tabular} & \begin{tabular}{@{}c@{}} 20.0\\ 82.5 \end{tabular} & \begin{tabular}{@{}c@{}} 0.0\\ 0.0 \end{tabular} & \begin{tabular}{@{}c@{}} 80.0 \\ 15.0\end{tabular} \\ \hline
16 & 32 & $1:2$ & $16\times 16$ & 0.669 & \begin{tabular}{@{}c@{}}0.377 \\ (0.344)\end{tabular} & \begin{tabular}{@{}c@{}} 0.06 \\ (0.05) \end{tabular} & \begin{tabular}{@{}c@{}} 0.23 \\ (0.19) \end{tabular} &\begin{tabular}{@{}c@{}}start \\ end\end{tabular} & \begin{tabular}{@{}c@{}} 0.0 \\ 5.0 \end{tabular}  & \begin{tabular}{@{}c@{}} 10.0 \\ 80.0 \end{tabular} & \begin{tabular}{@{}c@{}} 0.0 \\ 5.0 \end{tabular}  & \begin{tabular}{@{}c@{}} 90.0 \\  10.0\end{tabular}  \\ \hline 
32 & 32 & $1:1$ & $16\times 16$ & 0.637 & \begin{tabular}{@{}c@{}}0.203 \\ (0.219)\end{tabular} & \begin{tabular}{@{}c@{}} 0.04 \\ (0.03) \end{tabular} &  \begin{tabular}{@{}c@{}} 0.18 \\ (0.20) \end{tabular} &\begin{tabular}{@{}c@{}}start \\ end\end{tabular} & \begin{tabular}{@{}c@{}} 0.0 \\ 0.0 \end{tabular} & \begin{tabular}{@{}c@{}} 5.0\\ 40.0 \end{tabular}  & \begin{tabular}{@{}c@{}} 0.0\\ 55.0 \end{tabular}  & \begin{tabular}{@{}c@{}}95.0 \\5.0 \end{tabular}     \\ \hline 
20 & 50 & $2:5$ & $20\times 20$ & 0.684 & \begin{tabular}{@{}c@{}}0.453 \\ (0.480)\end{tabular} & \begin{tabular}{@{}c@{}} 0.03 \\ (0.02) \end{tabular} & \begin{tabular}{@{}c@{}} 0.11 \\ (0.10) \end{tabular} &\begin{tabular}{@{}c@{}}start \\ end\end{tabular} & \begin{tabular}{@{}c@{}} 0.0 \\0.0 \end{tabular}  & \begin{tabular}{@{}c@{}} 5.0\\ 90.0 \end{tabular} & \begin{tabular}{@{}c@{}} 0.0 \\ 0.0 \end{tabular} &  \begin{tabular}{@{}c@{}}95.0 \\ 10.0 \end{tabular}  \\ \hline 
25 & 50 & $1:2$ & $20\times 20$ & 0.687 & \begin{tabular}{@{}c@{}}0.323 \\ (0.350)\end{tabular} &\begin{tabular}{@{}c@{}} 0.02 \\ (0.02) \end{tabular} & \begin{tabular}{@{}c@{}} 0.14 \\ (0.10) \end{tabular} &\begin{tabular}{@{}c@{}}start \\ end\end{tabular} & \begin{tabular}{@{}c@{}}0.0 \\ 0.0 \end{tabular}  & \begin{tabular}{@{}c@{}}0.0\\ 55.0 \end{tabular}  & \begin{tabular}{@{}c@{}}0.0 \\ 45.0 \end{tabular}  & \begin{tabular}{@{}c@{}} 100.0\\ 0.0 \end{tabular}      \\ \hline
50  & 50 & $1:1$ & $20\times 20$ & 0.642 & \begin{tabular}{@{}c@{}}0.242 \\ (0.240)\end{tabular} & \begin{tabular}{@{}c@{}} 0.01 \\ (0.0) \end{tabular} & \begin{tabular}{@{}c@{}} 0.15 \\ (0.14) \end{tabular} &\begin{tabular}{@{}c@{}}start \\ end\end{tabular} & \begin{tabular}{@{}c@{}}0.0 \\0.0 \end{tabular}  &  \begin{tabular}{@{}c@{}} 15.0 \\ 20.0 \end{tabular} & \begin{tabular}{@{}c@{}} 0.0\\ 80.0 \end{tabular} &  \begin{tabular}{@{}c@{}} 85.0 \\ 0.0 \end{tabular}   \\ \hline 
\end{tabular}
\end{center}
\end{table*}

\begin{table*}[t]
\caption{Metrics (cf. Sec.~\ref{sec:metrics}) of 20~independent evolutionary runs with predefined zero predictions. Median values in brackets. \label{tab:measurementsEMPTY}}
\begin{center}
\begin{tabular}{|c|c|c|c|c||c|c|c||c|c|c|c|c|}
\hline
& & & \textbf{grid} & \textbf{median} & & \textbf{block} & \textbf{robot} & \multicolumn{1}{c|}{} &\multicolumn{4}{c|}{\textbf{structures}} \\
\cline {10-13}
\textbf{robots} & \textbf{blocks} & \textbf{ratio} & \textbf{size} & \textbf{fitness} & \textbf{similarity (\%)} & \textbf{movement} & \textbf{movement} &  \multicolumn{1}{c|}{} & \textbf{lines} & \textbf{pairs} & \textbf{clusters} & \textbf{dispersion} \\
\hline
10 & 32 & $5:16$ & $16\times 16$ & 0.930 & \begin{tabular}{@{}c@{}}0.773 \\ (0.812)\end{tabular}  & \begin{tabular}{@{}c@{}}0.0 \\ (0.0)\end{tabular} & \begin{tabular}{@{}c@{}}0.46 \\ (0.43)\end{tabular} &  \begin{tabular}{@{}c@{}}start \\ end\end{tabular} & \begin{tabular}{@{}c@{}} 0.0 \\ 5.0 \end{tabular} & \begin{tabular}{@{}c@{}} 10.0 \\ 15.0\end{tabular} & \begin{tabular}{@{}c@{}} 0.0 \\ 5.0\end{tabular} & \begin{tabular}{@{}c@{}} 90.0 \\ 75.0\end{tabular} \\ \hline
16 & 32 & $1:2$ & $16\times 16$ & 0.931 & \begin{tabular}{@{}c@{}}0.577 \\ (0.594)\end{tabular} & \begin{tabular}{@{}c@{}}0.0 \\ (0.0)\end{tabular} & \begin{tabular}{@{}c@{}}0.52 \\ (0.48)\end{tabular} &  \begin{tabular}{@{}c@{}}start \\ end\end{tabular} & \begin{tabular}{@{}c@{}} 0.0 \\ 0.0 \end{tabular}  & \begin{tabular}{@{}c@{}} 0.0 \\ 35.0 \end{tabular}  & \begin{tabular}{@{}c@{}} 0.0 \\ 15.0 \end{tabular}  & \begin{tabular}{@{}c@{}} 100.0 \\ 50.0 \end{tabular}  \\ \hline 
32 & 32 & $1:1$ & $16\times 16$ & 0.907 & \begin{tabular}{@{}c@{}}0.295\\ (0.296)\end{tabular} & \begin{tabular}{@{}c@{}}0.0 \\ (0.0)\end{tabular} & \begin{tabular}{@{}c@{}}0.42 \\ (0.40)\end{tabular} &\begin{tabular}{@{}c@{}}start\\ end\end{tabular} & \begin{tabular}{@{}c@{}} 0.0 \\ 7.5 \end{tabular} & \begin{tabular}{@{}c@{}} 5.0 \\ 40.0 \end{tabular}  & \begin{tabular}{@{}c@{}} 0.0 \\ 42.5 \end{tabular}  & \begin{tabular}{@{}c@{}} 95.0 \\ 10.0\end{tabular}     \\ \hline 
20 & 50 & $2:5$ & $20\times 20$ & 0.931 & \begin{tabular}{@{}c@{}}0.754 \\ (0.820)\end{tabular} & \begin{tabular}{@{}c@{}}0.0 \\ (0.0)\end{tabular} & \begin{tabular}{@{}c@{}}0.45 \\ (0.45)\end{tabular} &\begin{tabular}{@{}c@{}}start \\ end\end{tabular} & \begin{tabular}{@{}c@{}}0.0 \\ 2.5\end{tabular}  &  \begin{tabular}{@{}c@{}} 10.0\\ 2.5\end{tabular} & \begin{tabular}{@{}c@{}} 0.0\\ 5.0\end{tabular} &  \begin{tabular}{@{}c@{}} 90.0\\ 90.0 \end{tabular}  \\ \hline 
25 & 50 & $1:2$ & $20\times 20$ & 0.929 & \begin{tabular}{@{}c@{}}0.572 \\ (0.630)\end{tabular}  & \begin{tabular}{@{}c@{}}0.0 \\ (0.0)\end{tabular} & \begin{tabular}{@{}c@{}}0.52 \\ (0.46)\end{tabular} &\begin{tabular}{@{}c@{}}start \\ end\end{tabular} & \begin{tabular}{@{}c@{}} 0.0\\ 0.0 \end{tabular}  & \begin{tabular}{@{}c@{}} 10.0\\ 45.0 \end{tabular}  & \begin{tabular}{@{}c@{}}0.0 \\ 15.0 \end{tabular}  & \begin{tabular}{@{}c@{}} 90.0 \\ 40.0 \end{tabular}      \\ \hline 
50  & 50 & $1:1$ & $20\times 20$ & 0.907 & \begin{tabular}{@{}c@{}}0.351 \\ (0.360)\end{tabular} & \begin{tabular}{@{}c@{}}0.0 \\ (0.0)\end{tabular} &  \begin{tabular}{@{}c@{}}0.37 \\ (0.36)\end{tabular} &\begin{tabular}{@{}c@{}}start \\ end\end{tabular} & \begin{tabular}{@{}c@{}} 0.0 \\ 5.0\end{tabular}  &  \begin{tabular}{@{}c@{}} 0.0 \\ 45.0 \end{tabular} & \begin{tabular}{@{}c@{}}  0.0 \\ 35.0 \end{tabular} &  \begin{tabular}{@{}c@{}} 100.0 \\ 15.0  \end{tabular}   \\ \hline 
\end{tabular}
\end{center}
\end{table*}

To bias the emergence towards desired structures, we predefine sensor predictions in the next step (cf.~\cite{kaiser18}). 
We set the robot sensor predictions to~0 (i.e., $s_0 = s_1 = s_2 = s_3 = s_4 = s_5 =0$) and vary the block sensor predictions in three different scenarios. 

First, we aim for pairs and lines by predefining the sensor predictions for the two block sensors in front of the robot to~1 (i.e., $s_6 = s_9 = 1$) while all other predictions are set to~0 (i.e., $s_7 = s_8 = s_{10} = s_{11} = 0$). 
We require all robots to have a pair or line of blocks directly in front of them to maximize fitness. 

The mean best fitness is about $80\%$ in all experiments, cf.~Tab.~\ref{tab:measurementsPREL}, that is up to $8.4$~pp lower than in our first experiments.
Fig.~\ref{fig:fitnessPRE} shows the increase of best fitness over generations. 
It is representative for the fitness curves observed in all following experiments. 

\begin{figure}[t]
    \centering
     \subfloat[initial position]{\includegraphics[width=0.44\linewidth]{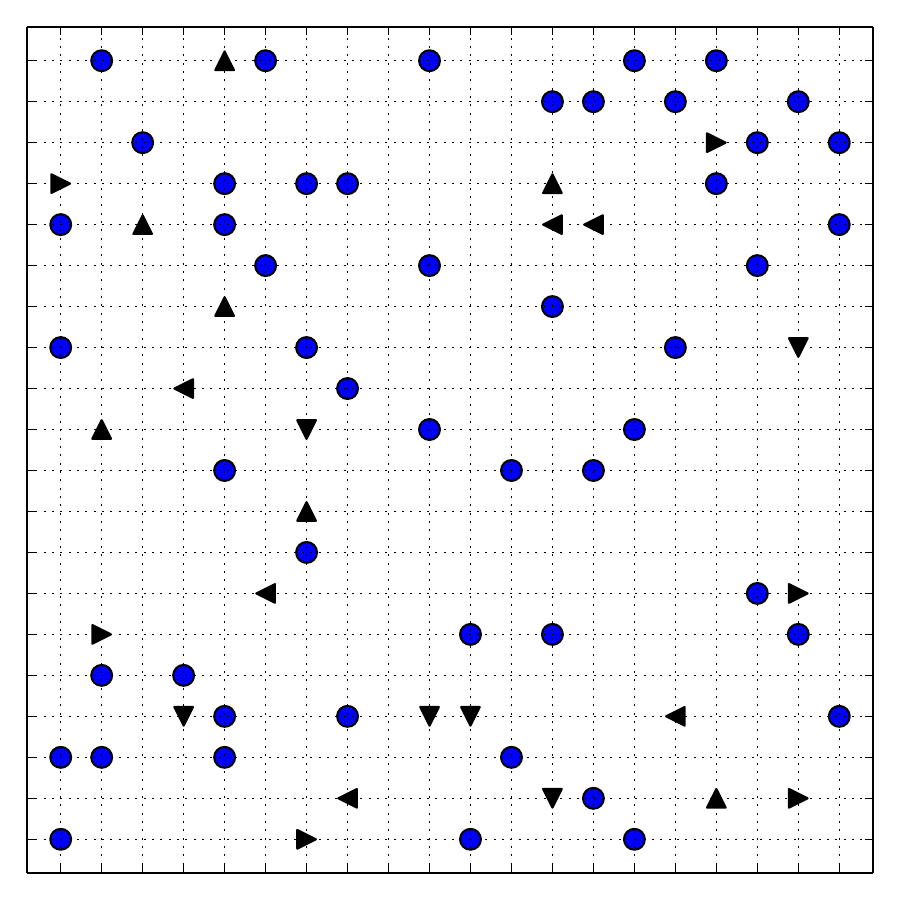}}
     \subfloat[final position]{\includegraphics[width=0.44\linewidth]{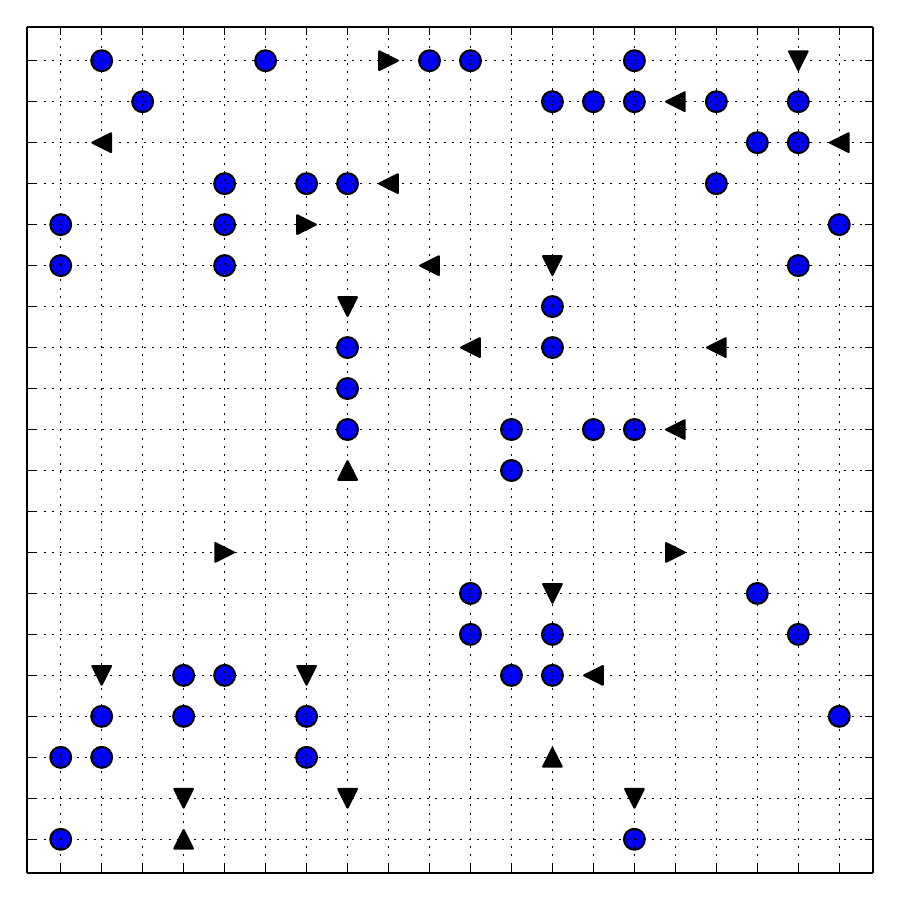}}
    \caption{Resulting pair structure with predefined sensor predictions using 25~robots and 50~blocks on a $20\times 20$ grid. Robots are represented by triangles, blocks by circles. Triangles give the robots' headings.}
\label{fig:predefined}
\end{figure}

We find that block positions were altered in all runs. 
The median similarity decreased by roughly $20$~pp on the smaller grid and by $30$~pp on the larger grid for the lower robot-to-block ratios compared to our initial experiments. 
Thus, we find a greater alteration of the structures using low robot-to-block ratios than before. 
This finding is supported by the formed structures. 
The robots decrease dispersion during the runs by $67.5$~pp. 
The majority of formed structures are pairs (Fig.~\ref{fig:predefined}) and we observe lines.

Next, we want to provoke that the robot swarm forms clusters of blocks. 
We predefine that all block sensors are predicted to see a block  (i.e., $s_6 = s_7 = s_8 = s_9 = s_{10} = s_{11}=1$). 
The median best fitness is $63\% $ to $69\%$ and thus, around $20$~pp lower than in our previous experiments. 
We infer that the task complexity increased. 
In all runs, block positions were altered. 
The similarity of block positions decreases with the swarm density on both grids from around $50\%$ to $22\%$. 
Compared with all previous experiments, we reach the lowest similarity in this scenario and thus, the most intense pushing of blocks. 
As before, the percentage of dispersion structures decreases by $65$ to $100$~pp during the runs.  
We observe that the amount of emerging clusters varies with swarm density. 
Mainly pairs emerge for the two lower swarm densities on both grids. 
While no clusters form for the lowest swarm density, one run on the smaller grid and roughly half of the runs on the larger grid result in clusters for the intermediate density.

\begin{figure}[t]
    \centering
     \subfloat[initial position]{\includegraphics[width=0.44\linewidth]{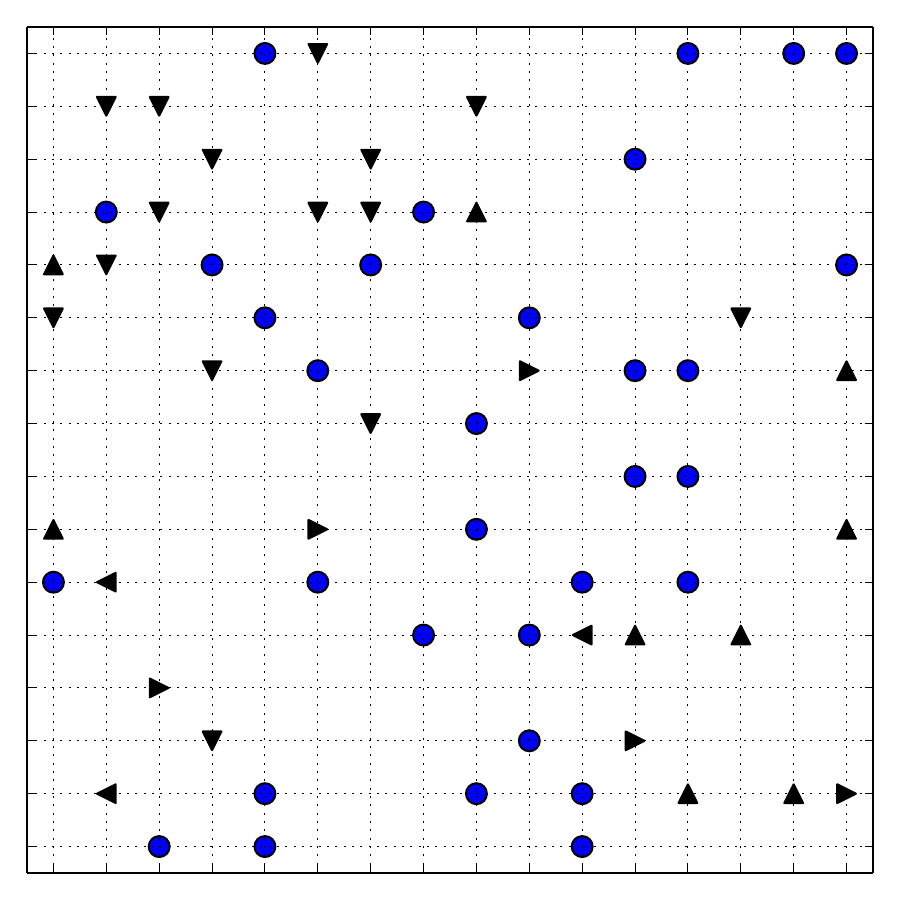}}
     \subfloat[final position]{\includegraphics[width=0.44\linewidth]{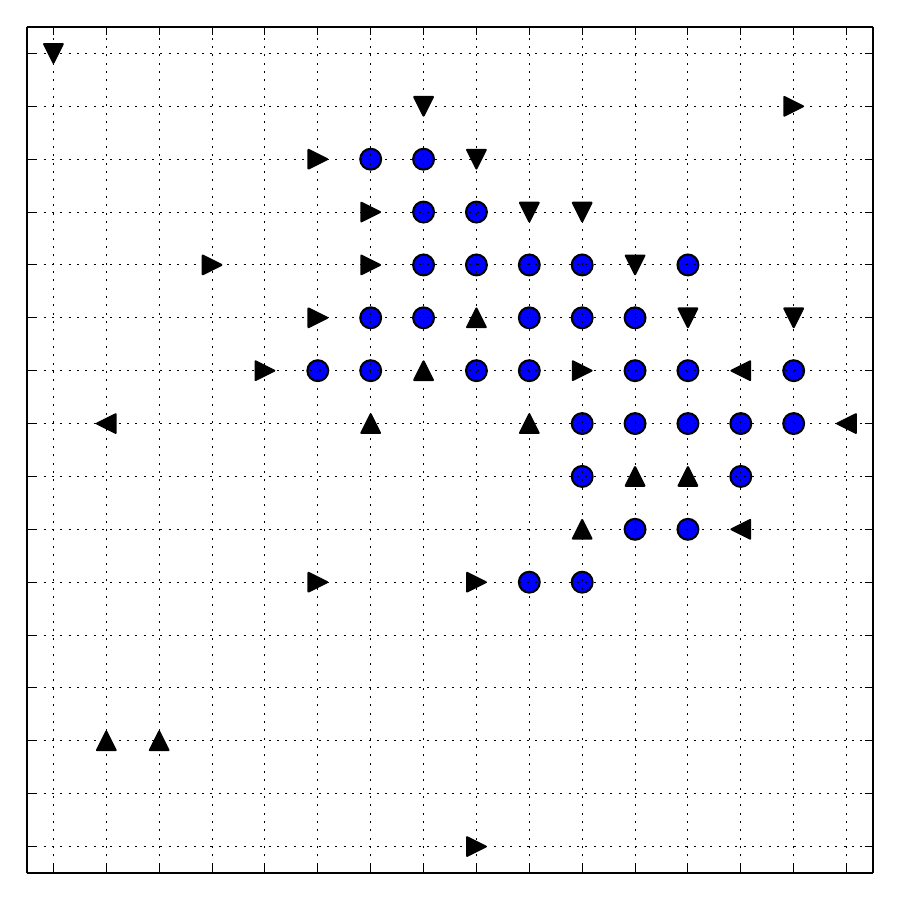}}
    \caption{Resulting cluster with predefined sensor predictions using 32~agents and 32~blocks on a $16\times 16$ grid. Robots are represented by triangles, blocks by circles. Triangles give the robots' headings.}
\label{fig:predefinedCluster}
\end{figure}

The runs with the highest swarm density and a $1:1$ robot-to-block ratio mainly result in clustering on both grid sizes (Fig.~\ref{fig:predefinedCluster}) but the number is $25$~pp higher on the larger grid.
We conclude that the task is harder for smaller grids and lower swarm densities. 
There is more robot and block movement in the last $\tau$~time steps for the lower grid size. An increased runtime may lead to more clusters. 

In a last scenario, we predefine that robots predict no blocks in front of them (i.e., $s_6 = s_7 = s_8 = s_9 = s_{10} = s_{11}=0$). 
We observe that blocks are moved around in almost all runs except for one run on the larger grid and for two runs on the smaller grid with the lowest swarm density. 
The median best fitness is above $90\%$ for all runs and between $1.5$ and $6.3$~pp higher than for the runs without predefined predictions. 

We observe various structures. 
Robots try to disperse themselves to neither perceive robots nor blocks. 
Grouping blocks in clusters, pairs, or lines may be beneficial but especially for lower swarm densities the initial structures may already allow all robots to disperse and find an isolated position. 

In summary, we can engineer self-organized construction by predefining sensor value predictions. 
Some structures, such as clusters or pairs, can easily be engineered while predefining not to sense any blocks provides rather an additional intrinsic driver to group blocks. 

\section{Discussion and Conclusion}

Our study on self-organized swarm construction with minimal surprise shows that our approach can be used to evolve swarm behaviors which require interaction with the environment, too. 
We are able to engineer swarm construction behaviors by predefining sensor predictions. 

In future work, we want to investigate the influence of block density, swarm density and grid size more thoroughly. 
First investigations with discrete sensors indicating either block, robot, or empty cell showed that it complicates the problem. 
This may require a study of the ANN parameters.
Preliminary investigations with seeds of block formations placed initially in the environment showed that they can effectively trigger the grouping of blocks at designated spots.
However, seeds were not as effective as predefining predictions.
We want to investigate more seed shapes and the combination with predefined predictions in future. 

Overall, we are confident to extend our presented results in future work to evolve more complex and interesting swarm construction behaviors using minimal surprise. 
The step from simulation to robot hardware seems feasible.

\bibliography{ref}

\end{document}